\documentclass[11pt,a4paper]{article}

\usepackage[hyperref]{acl2021}
\usepackage{times}
\usepackage[disable]{todonotes}
\usepackage{latexsym}
\usepackage{booktabs}

\usepackage{microtype}
\usepackage{mdframed}
\usepackage{multirow}
\usepackage{multicol}
\usepackage[ruled,vlined]{algorithm2e}
\usepackage{algorithmic}
\usepackage{graphicx}
\usepackage{subcaption}

\newcommand{\ndb}{{NLDB}}
\newcommand{\ndbs}{{NLDBs}}
\newcommand{\wikidb}{{\textsc{WikiNLDB}}}
\newcommand{\example}[1]{``\emph{#1}''}
\newcommand{\triple}[1]{{\tt #1}}
\newcommand{\tfidf}{{TF$\cdot$IDF}}

\title{Database Reasoning Over Text}

\author{James Thorne \\
  University of Cambridge \\ 
  {\tt jt719@cam.ac.uk} \\\And
  Majid Yazdani \\
  Facebook AI \\
  {\tt myazdani@fb.com} \\\And 
 Marzieh Saeidi \\
 Facebook AI \\
{\tt marzieh@fb.com} \\ \AND

Fabrizio Silvestri\\
Sapienza University, Rome \\
{\tt fsilvestri}\\{\tt@diag.uniroma1.it} \\ \And

Sebastian Riedel \\
Facebook AI \\
University College London \\ 
{\tt sriedel@fb.com} \\ \And

Alon Halevy \\
Facebook AI \\
{\tt ayh@fb.com}
}

\aclfinalcopy
\date{}

\begin{document}
\maketitle
\begin{abstract}
Neural models have shown impressive performance gains in answering queries from natural language text. 
However, existing works are unable to support database queries, such as \example{List/Count all female athletes who were born in 20th century}, which require reasoning over {\em sets} of relevant facts with operations such as join, filtering and aggregation. We show that while state-of-the-art transformer models perform very well for small databases, they exhibit limitations in processing noisy data, numerical operations, and queries that aggregate facts.
We propose a modular architecture to answer these database-style queries over multiple spans from text and aggregating these at scale. We  evaluate the architecture using  \wikidb,\footnote{\url{https://github.com/facebookresearch/NeuralDB}} a novel dataset for exploring such queries. Our architecture scales to databases containing thousands of facts whereas contemporary models are limited by how many facts can be encoded.
In direct comparison on small databases, our approach increases overall answer accuracy from 85\% to 90\%. On larger databases, our approach retains its accuracy whereas transformer baselines could not encode the context.




\end{abstract}

\section{Introduction}


\begin{figure}[t]
    \centering
    \includegraphics[width=\linewidth]{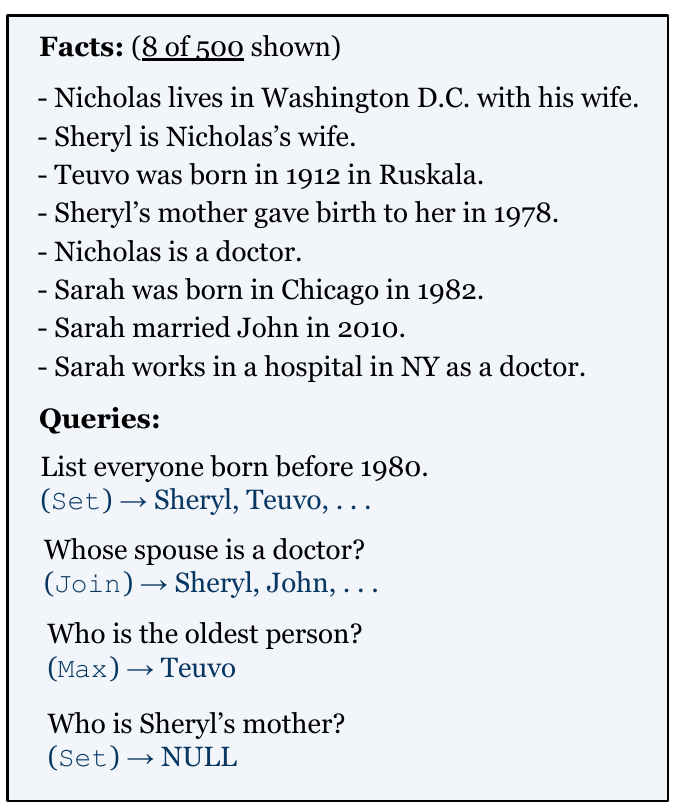}
    \caption{Examples of set and aggregation queries over a natural language database: a database where facts are stored in free-form text without the need for a schema.}
    \label{fig:actual_data}
\end{figure}

Question answering (QA) over text has made significant strides in recent years owing to the availability of new datasets and models. 
Machines have surpassed human performance on the well-known SQUaD task~\citep{Rajpurkar2016} where models extract answer spans from a short passage of text. The subsequent body of work has further considered incorporating retrieval from large corpora such as Wikipedia~\cite{dhingra2017quasar, joshi2017triviaqa, kwiatkowski2019natural} to identify relevant information, conditioning answer generation~\cite{chen2017reading, lewis2020retrieval, Izacard2020}. More sophisticated architectures have been proposed with incremental retrieval for multi-hop QA~\cite{xiong2020answering, das2019multi}, where several passages are required, which may have low lexical or semantic similarity with the question.

This paper considers the problem of answering questions similar to database queries, such as those shown in Figure~\ref{fig:actual_data}. 
For example, the query \example{List all the female athletes in Wikipedia who were born in the 20th century}, requires reasoning over hundreds or thousands of facts, retrieved from multiple Wikipedia pages, and applying set-based filters to them (e.g., gender, birth date). If our query further asked how many such athletes exist, we would have to perform an aggregation function to count the result set. 
The ability to answer the aforementioned queries would enable a new kind of database \citep{thorne2021_VLDB} where facts can be described in natural language and would therefore obviate the need for a pre-defined schema, which is a major limitation of current database systems. An example application for such flexible text databases exists in the area of storing knowledge for personal assistants 
where users store data about their habits and experiences, their friends and their preferences, for which designing a schema is impractical. 

We introduce \wikidb, a benchmark dataset for exploring database reasoning over facts expressed in natural language. \wikidb\ contains a number of query types that 
require systems to return large set-based answers and aggregate over these (with operators such as \texttt{count}, \texttt{min}, and \texttt{max}). Our dataset is generated using publicly available knowledge graph data, enabling large volumes of instances to be generated with minimal effort. Most queries in \wikidb\ require reasoning over hundreds of facts to generate answers, exposing limitations in current neural models.
In contrast to DROP~\citep{Dua2019} where queries are answered over \emph{single} passages, and bAbI~\cite{weston2015towards}, where each query is based on a context of less than 20 facts, our dataset scales from databases of 25 instances to 1000, and could be extended further.


 
We also introduce a modular architecture to support database reasoning over text and characterize its behavior on our reference dataset. We find that even on small databases of 25 facts, naive application of transformers is insufficient. When provided with only the relevant facts, the baseline yields an answer accuracy of 85\%, whereas applying our proposed architecture yields 90\% by better answering queries, such as {\tt count}, that require computation. 
It is well known that transformer models do not scale well to large inputs due to the use of self-attention. We found that mechanisms such as Fusion in Decoder \citep[FiD]{Izacard2020} and LongFormer~\citep{Beltagy2020}, which mitigate the scaling issue, harm the model: combining more than 2 facts with FiD resulted in answer accuracies of 76\% and 39\%, respectively.
These issues were mitigated by our approach which generates intermediate query-based derivations of small numbers of facts in the database, before using conventional computation to aggregate the results.


\section{Answering Database Queries over Text}

\subsection{Problem Definition}
We refer to corpora that consist of unordered collections of facts expressed as short natural language sentences as Natural Language Databases (\ndb s). 
For example, a corpus may include all the utterances given to a personal assistant by its user, or all the claims uttered by a political figure. The texts in our corpora are similar to databases as they are sets of stand-alone facts. But unlike a database, they are not expressed as rows or triples in a pre-defined schema. For example, a sentence containing a single fact, \example{Gustavo likes espresso} 
or multiple facts, such as \example{Robertson Howard, who attended the University of Virginia, is buried in the Congressional Cemetery}.


A query $Q$ over a database, $D$, produces a set of answers: $Q(D) = \{a_1,\ldots,a_l\}$.
We consider the following four query types (see examples in Table~\ref{tab:db-examples}):
(1) {\em Set queries} are extractive queries that return a list of spans, such as entities, from the facts.
(2) {\em Boolean queries} return a True/False answer.  
(3) {\em Aggregation queries} require computation over answer sets with an operator, such as {\tt count}, {\tt min} and {\tt max}. For example: \example{How many people work for Yale Law School?}).
(4) {\em Join queries} require the combination of two (or more) facts to produce each answer. We combine join operations with set, Boolean and aggregation queries.
For example, the query \example{Who works in a company in France?} considers both the relationship between people and employer as well as company locations.





\subsection{Challenges}

\begin{figure*}[ht]
    \centering
    \includegraphics[width=1\linewidth]{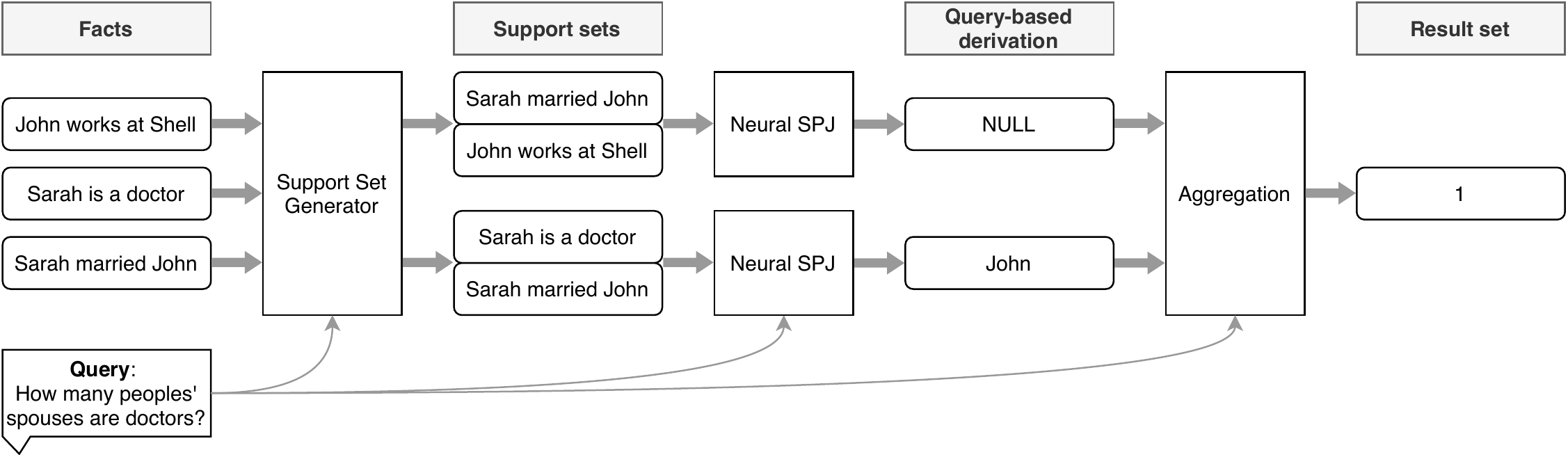}
    \caption{Overview of the proposed architecture. Consisting of a support set generator, SPJ and aggregation}
    \label{fig:overview}
\end{figure*}

The NLP treatment of question answering, where systems encode the query and context (containing the background knowledge), forms a good starting point for \ndbs. Common model architectures are based on the transformer \citep{Vaswani2017} in an encoder-decoder configuration. The encoder uses self-attention to conditionally encode the context with the query and the decoder allows conditional generation of outputs that are not necessarily present in the input. 
To scale question answering to reason over large knowledge-sources such as Wikipedia, task formulations typically retrieve text-spans from a corpus to condition answer generation \citep{chen2017reading, dhingra2017quasar}. 
However, several challenges encountered in \ndbs{} preclude direct application of these techniques:

\paragraph{Scale} To scale neural reasoning to databases of non-trivial size, it would not be feasible to encode the entire database as input to the transformer. 
Question answering systems combine a retrieval mechanism to select relevant spans from knowledge sources as context. This task is usually referred to as open-domain QA~\cite{Lewis2020,Izacard2020}. 
It is common to use a maximum input size of $512$ or $1024$ tokens for context. While extensions such as Linformer \citep{Wang2020}, Longformer \citep{Beltagy2020} and Fusion in Decoder \citep{Izacard2020} enable larger contexts to be encoded, their application of self-attention varies and the number of tokens that may be encoded is limited by GPU memory. 

\paragraph{Multiple answer spans} The NLP formulation of question answering typically requires extracting a span from a single document or generating a short answer. Answering queries in a \ndb\ may require processing a large number of facts, generating a large number of items as answer, hundreds or thousands, and performing aggregations over large sets.

\paragraph{Locality and document structure} \ndb s do not enjoy the locality properties that usually hold in open-domain QA. In \ndb s, a query may be dependent on multiple facts that can be anywhere in the database.  In fact, by definition, the current facts in a database can be reordered and the query answers should not change. In contrast, in open-domain QA, the fact needed to answer a given question is typically located in a paragraph or document with multiple sentences about the same subject, in combination with a document title, where this additional context may help information recall. 

\paragraph{Conditional retrieval} 
Similar to open-domain question answering, \ndbs{} mandate an information retrieval component. 
When determining which facts to input to the model, \ndb s may require conditional retrieval from the database. For example, to answer the query \example{Whose spouse is a doctor?} we'd first need to fetch spouses and then their professions. Recent work on multi-hop query answering (e.g.,~\citet{asai2019learning}), has started considering this issue but is restricted to the case where we're looking for a single answer. In \ndb s, we may need to perform multi-hops for sets of facts.  

\section{Architecture for querying \ndbs}

\label{section:architecture}
To address the aforementioned challenges, we propose an instance of a Neural Database architecture \citep{thorne2021_VLDB} that operates over textual facts with parallelizable non-blocking operators before aggregating the results. The three core components of the architecture, shown in Figure~\ref{fig:overview}, are a Support Set Generator (SSG) which retrieves small sets of relevant facts called support sets,  a parallelizable non-blocking Select-Project-Join (SPJ) operator which generates intermediate answers that can be unioned to produce the final answer, and an optional aggregation stage which uses conventional computation to perform numerical reasoning. 
The key insight underlying our architecture is to leverage neural models for what they excel at, namely, reasoning over a small set of facts. 

\paragraph{Neural SPJ Operator} Given a single support set and a query, the SPJ (Select-Project-Join) operator outputs a machine readable intermediate representation of the answer that can be generated from the support set. For example, given the query \example{Who was born in Montevideo?} and the support set \{\example{Mario Sagario was born in Montevideo, Uruguay, ...}\}, the Neural SPJ would output the entity literal \triple{Mario Sagario}. 
Examples of outputs are provided in Figure~\ref{fig:query_derivations}.

The SPJ operator is performing three functions: (1) for support sets that are insufficient to answer a question, the operator should return no output; (2) for queries that require short chains of reasoning over multiple facts, the SPJ operator joins the facts when generating the output; and (3) the SPJ generates a projection of the support set to a machine readable format dependent on the given query, and whether computation or aggregation is required. 

Because the SPJ operator is run in parallel, it can scale independently of the limitations on the size of the input of a single transformer. In contrast, the use of self-attention when encoding all facts as one input precludes parallelization, has high latency, and is limited by the memory required to compute the self-attention.
By using the SPJ operator to perform query-dependent information extraction, aggregations can be performed over the generated outputs using conventional computation, which trivially scales to thousands of operands. Furthermore, this allows large result sets to be generated by the model, whereas accurately decoding long sequences using an encoder-decoder architecture remains an open challenge \citep{Hupkes2020}.

\paragraph{Support Set Generator (SSG)} A support set contains the minimal subset of sentences from the database needed to generate one single operand for the aggregation module by the SPJ operator. For example, for queries that are answered by a single sentence, e.g., \example{Who is Sheryl's husband?}, the support set containing a single fact should be returned, e.g.,  \{\example{Sheryl is Nicholas's spouse}\}. 
The output of the support set generator is a \emph{set} of support sets, each of which is fed independently to a downstream SPJ module. 
Support sets may not be pairwise disjoint because some facts may be required for multiple answers. 

The SSG output should satisfy the following two properties:
(1) If multiple facts are needed to produce an intermediate answer, they should all be in the support set.  For example, if we queried \example{When was Sheryl's husband born?}, the support set should include a fact stating who the spouse is and a fact describing when they were born.   
(2) When performing aggregation, or outputting a set of answers, multiple support sets must be generated, each containing enough information to generate the intermediate results that are aggregated.
For example, for the query \example{Who is the oldest person?}, each of the support sets would independently contain a fact that includes a person and indicates their age. 

\paragraph{Aggregation}
The outputs of the SPJ modules are intermediate answers to the query. For some queries, e.g., \example{who lives in London?}, the final answer is simply the union of the intermediate answers. In other cases, e.g., \example{how many countries grow coffee?}, an aggregation operator needs to be applied to the union of intermediate answers. Because output of the SPJ operators are machine readable, we can hence guarantee accuracy and scalability by performing aggregation using conventional computation. In this paper, we consider the aggregation functions  \triple{min}, \triple{max} and \triple{count}.


\section{The \wikidb\ dataset} 
\label{section:data}

In this section we introduce \wikidb, a novel dataset for training \ndbs{} which is generated by transforming structured data from Wikidata~\cite{vrandevcic2014wikidata} into natural language facts and queries.
Wikidata stores triples of the form \triple{(S,R,O)}, where $R$ is a relationship between the subject $S$ and the object $O$, e.g., \triple{(Tim Cook, employedBy, Apple)}.
The scale and breadth of Wikidata enables us to generate databases of many sizes and variety. 
  
\paragraph{Facts} To automate generation of questions and answers, sentences must be grounded in Wikidata identifiers. 
One approach to generate facts would be to use templates or collect them through grounded information extraction datasets such as T-REx \citep{elsahar-etal-2018-rex}. However, to ensure wider linguistic variety as well as accuracy of the mapping, we use verbalizations of knowledge graph triples that are synthesized through a sequence to sequence model. 
Concretely, we use generated sentences from KELM \citep{agarwal2020large}, which are not grounded with Wikidata IDs, and generate a post-hoc mapping back to Wikidata.
For example, given the sentence: \example{The Slice of Life manga series The Film Lives On was written by Osamu Tezuka.} we map it to the Wikidata triple \triple{(Q11332517,P50,Q193300)}. 
Our mapping is a two-step process: firstly, we look up entity names from Wikipedia, returning multiple matches for \triple{Osamu Tezuka}, and secondly filter these based on which have an author relations to \triple{The Slice of Life} in the Wikidata graph.
While out of scope for this paper, this technique could be applied to generate training datasets for novel domains.
\wikidb\ uses both atomic facts in KELM (about a single relation of an entity) or composite facts (about multiple relations). 

\paragraph{Queries} Following previous work on large-scale question answering \citep{hartmann-marx-soru-2018, talmor2018web}, queries are generated using templates.
For each relation and operator, multiple templates were written by the authors where placeholders can be replaced with the subject and objects for each relation. 
While multiple templates are used to ensure variety, these are limited in diversity in comparison to the facts.
Templates were generated for the first 25 relations on Wikidata with mapped data in KELM. 
To generate queries that require joins we apply the same technique, combining to combine two or more connected relations, chaining the entities. We further select the 15 most popular relations and generate additional templates which chain the two relations. For example, we chain \triple{(Y,locatedIn,Z)} and \triple{(X,employedBy,Y)} to create a template for the query \example{Does \$X work at a company based in \$Z?}.


\paragraph{Data Quality} We manually inspect randomly selected queries and facts and score them using the categories introduced in this section.
For queries, we sample 70 instances, 10 for each query type. We score each query for fluency and intelligibility. Out of 70 queries, only one question was marked as non-fluent due to a typo which was corrected for the final dataset. All 70 queries were intelligible. We observed that the clarity of some queries depended on the facts in the database to provide context (e.g. ``Who is male?''), but otherwise met the task requirements.

To assess the quality of mapped facts from KELM, a sample of 50 was evaluated based on 6 categories: intelligibility, fluency, 
inclusivity (conveying information from all the mapped relations), faithfulness to these relations, and whether extraneous information (not in the mapped relations) is present. $49/50$ facts were intelligible and $45/50$ facts were fluent. The remaining $5$ had redundant information or missing conjunctions. $50/50$ facts contained all mapped relations and $48/50$ were faithful to these relations. $8/50$ facts had extraneous information for relations that could not be mapped. 
The relations that could not be mapped are not used for query generation and did not affect how answers were automatically generated. 

\paragraph{\wikidb \ Statistics}
We create databases over 25 common relationships from Wikidata, and create $643$ templates from which queries are phrased.  For \emph{join}-type queries, we chain a further $15$ relations with a further $86$ template fragments. The relations we chose were selected from a weighted sample of the most common entity types in KELM.
In total, we generate five variants of the dataset containing databases of size 25 to 1000 facts where each fact has between 30-50 tokens.
Dataset statistics are reported in Table~\ref{tab:data_stat}. 

\begin{table}[]
\centering
\small
\begin{tabular}{@{}ccccc@{}}
\toprule
\multirow{1}{*}{\textbf{DB Size}}  &
\multirow{2}{*}{\textbf{Avg $\#$Q/DB}}  &
\multicolumn{3}{c}{\textbf{$\#$DBs}} \\ \cmidrule(l){3-5}  

 \multirow{1}{*}{(up to)} &  & \textbf{Train} & \textbf{Valid} & \textbf{Test} \\ \midrule
25 & 8 & 4000 & 631 & 621   \\
50 & 7 & 4986 & 498 & 499 \\
100 & 13 & 2500 & 250 & 250 \\
250 & 53 & 1000 &  100 & 100  \\
500 & 66 & 500 & 50 & 50  \\
1000 & 70 & 250 & 25 & 25  \\
\bottomrule
\end{tabular}
\caption{The statistics for datasets with varying size of DBs (i.e. number of facts). Average number of queries per each DB instance and also the number of DB instances per split is displayed. }
\label{tab:data_stat}
\end{table}



\section{Models}
\subsection{Neural Select-Project-Join}
\begin{figure}
    \centering
    \includegraphics[width=0.9\linewidth]{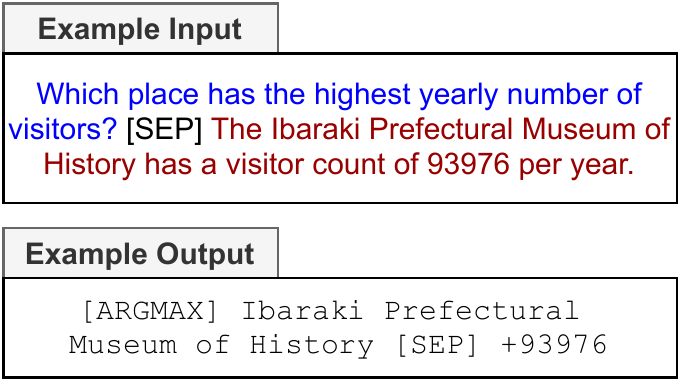}
    \caption{Example input and output of the Neural SPJ operator (blue: query, brown: support set sentences)}\label{fig:query_derivations}
\end{figure}

The SPJ operator is trained as a sequence-to-sequence model to generate intermediate results from a support set and a given query.  All facts in the support set are concatenated with the query before being input to a transformer model. 

The model is trained to output different derivations depending on the query type. For the \texttt{min}, \texttt{max} operators, the projection is a machine-readable key-value pair, illustrated in Figure~\ref{fig:query_derivations}. For example \example{which place has the highest yearly number of visitors?} has the projection of the form: \triple{(place, number of visitors)} allowing an argmax  operation by the downstream aggregation module. For queries with Boolean answers, the output is a token indicating whether the answer is true or false. And for all other queries where a set of results is returned or counted, the output is simply a span, such as an entity or numerical value, extracted from the support set. 

Even though we use intermediary annotation for the SPJ operator, we believe that collecting such annotation is a simpler labeling task compare to collecting the answers to the queries. For example, given the fact \example{Serena Jameka Williams (born September 26, 1981) is an American professional tennis player and former world No.} and the query \example{List all the female athletes who were born in 20th centure.}, it seems relatively simple to provide the label \example{Serena Jameka Williams}. However, it is non-trivial to produce a list of potentially hundreds of entities as answer (e.g. [\example{Serena Jameka Williams, Simona Halep, Mary Lou Retton, Megan Rapinoe, Kim Simmone, Mary Abichi, $\ldots$}]). The training of the components in our proposed architecture does not depend on the final answer and instead, on the simpler intermediary labels.

\paragraph{Predicting Aggregation Operator}
Rather than using a separate classifier to predict the question type, we encode the choice of operator as a special token that is predicted by the SPJ operator prepended to the model output (Figure~\ref{fig:query_derivations}).  The aggregation operator is chosen using a majority vote over all generated derivations from all support sets.

\paragraph{Negative Example Generation}
It is important for the SPJ to be resilient to extraneous facts that might be returned by a low-precision high-recall SSG.
Negative instances for training are generated in two ways: (1) queries are paired with randomly sampled facts and the model is trained to generate a \texttt{NULL} projection (indicating the support set does not contribute to the answer). For example, a fact about someone's date of birth isn't useful when answering a query about the visitor count of an attraction. (2) for a portion of the training instances, we additionally sample extraneous unrelated facts and append these to the support sets simulating false-positive facts from the SSG.  

\subsection{Support Set Generator}
 \label{section:architecture:support-set-generation}
For simple queries over single facts, conventional information retrieval, such as \tfidf\ 
could be considered a primitive SSG. 
However, this would not scale for joins, aggregation queries or for queries outputting a set of answers as generating relevant sets requires incremental decoding, conditioning on already retrieved facts. 

Naively generating the set of all relevant support sets, $SSG_Q(D) \subset \mathcal{P}(D)$, would be intractable as it is akin to enumerating the powerset. We construct support sets efficiently by taking an incremental approach, starting from the empty set (see Algorithm~\ref{ssg-alg}). 
At each step, the classifier considers the partially generated support set $\hat{D}_k$ and the query and predicts which candidate facts $u_i \in D$ from the database should be added, or whether to stop the iteration, these choices being modeled as a multi-label classification task.
If {\tt STOP} is predicted, the partial result set $\hat{D}_k$ is closed (i.e., it forms part of the output); otherwise, for each fact added, a new intermediate ($open$) support set is generated which is explored in the next iteration.
For efficiency, we use a bi-encoder architecture that independently encodes the facts in the database and the state (query and a partial support set) and computes the inner product between the encoded representations to generate a score: $C_U(u_i)^T C_V(Q,\hat{D}_k)$. The encoders are pre-trained transformers fine-tuned to yield a high inner product between the state's encodings and relevant facts to be added.  At prediction time, the vectors encoding the facts are static and are pre-computed offline.
At each step, $t$, we encode the state using a transformer by concatenating the query tokens and the facts in the partially generated support set $D_k$.
The SSG is trained with full supervision of all partial support sets from the dataset and trained to predict which facts to add to the support set using a contrastive loss.

\begin{algorithm}[t]
\small
 \SetKw{In}{in}
 \KwIn{Bi-encoders $C$: $C_U$ (for actions), $C_V$ (for state), Database $D$, Query $Q$, Threshold $\tau$}
 \KwOut{Set of support sets ($\hat{D}_1,\ldots,\hat{D}_b$) $\subset \mathcal{P}(D)$ }
 open := \{\{\}\}\, closed := \{\}\; 
 U := $\lbrack C_U(u_1); \ldots; C_U(u_n); C_U(\mathtt{STOP}) \rbrack$ for $u_i \in D$\;
 
 \While{open $\neq$ \{\}}{
  next := \{\}\;
  
  \For{$\hat{D}_k$ \In open}{ 
    V := $\lbrack C_V(Q, u_1 \ldots u_m)\rbrack$, for $u_i \in \hat{D}_k$\;
    
    A := MIPS($U$,$V$,$\tau$)\;
    \For {$a_j$ \In $A$} {
        \If{$a_j$ == \texttt{STOP}} {
            $closed := closed$ $\cup \{\hat{D}_k\}$;
        } 
        \Else{
            $next := next$ $\cup \{ \{ a_j \cup \hat{D}_k\}\}$\;
        }
        
    }
          
    }
    
  open := next\;
 }
 \Return closed;
 \caption{SSG modeled as multi-label classification: using maximum inner product search (MIPS) over vector encodings of facts $U$ and state $V$ }
 \label{ssg-alg}
\end{algorithm}

\paragraph{Complexity of SSG} The inner loop of Algorithm~\ref{ssg-alg} involves a Maximum Inner Product Search (MIPS) between the encoded state and the encodings of the facts, which is linear in the number of facts. Approximate search, such as FAISS~\cite{Johnson2019}, accelerate retrieval to $O(\log^2 n)$. If we assume a query needs a maximum of $b$ support sets, and the average size of a support set is $m$, then the complexity of the SSG algorithm is
$O(bm\log^2 n)$. Both $b$ and $m$ are bounded by the number of facts in the database $n$, but in practice we'd expect only one of $b$ or $m$ factors to be large. However, there is fertile ground for developing methods for indexing (and/or clustering) the facts in the database so that only few facts need to be considered in each iteration of the inner loop of the algorithm, leading to significant speedups.

\subsection{Baselines}
We compare our proposed architecture to transformer-based models that explore the effect of three attention mechanisms representative of the state-of-the-art.
Self-attention in transformers captures both \textit{inter}-fact as well as \textit{intra}-fact interactions between tokens. However, computing self-attention is quadratic with respect to memory and scaling beyond 1024 tokens is non-trivial.
In our baselines, the task formulation is a sequence to sequence model, similar to that used in question answering. All (relevant) facts are encoded with the query and the transformer is trained to predict the answer without using any intermediate representations.
We compare \textit{full} self-attention against independently encoding the facts (in the context of the query) and fusing the embeddings in the decoder \citep[\textit{Fusion in Decoder (FiD)}]{Izacard2020}. Because FiD independently encodes contexts, run-time complexity is reduced to be linear with respect to the number of facts at the expense of not having inter-fact attention. 
We additionally compare to using windowed attention over facts with global attention to the query using \textit{Longformer} \citep{Beltagy2020}. Inter-fact attention is captured only within the window.

\section{Implementation}
We use the HuggingFace \cite{wolf2020huggingfaces} transformers library and its implementations of T5 and Longformer.  
For SSG, we use BERT to generate encodings, which has a comparable architecture to T5. 
The learning-rate for fine-tuning and number of epochs were selected through maximizing the Exact-Match (EM) accuracy on a held-out validation set for the tasks. For each experiment, we train 3 separate models with different seeds and report mean accuracy. 
The SPJ models are only trained on the small database of $25$ facts and applied to larger databases at test time.

For most queries, we measure correctness using Exact Match (EM), which is~1 if the answer string generated by the model is exactly equal to the reference answer and~0 otherwise.  This metric is used to score outputs where either a Boolean, null answer, string or numeric answer is expected. When a set of results is returned, we compute the $F_1$ score considering exact matches of set elements. When comparing models and reporting results, we report macro-averages over all instances in the test set. We collectively refer to this as \textit{Answer Accuracy}.

\section{Experiments \& Results}

\label{sec:wholedb}
We first consider the suitability of transformer models over small databases of 25 facts comparing two information retrieval settings: 
\emph{PerfectIR}, which is representative of other question answering approaches that combine an information retrieval system to select only the facts needed to answer a query, and \emph{WholeDB}, where the entire database is encoded by the model, assessing resilience to unrelated information and noise. 

\begin{table}
\centering
\small
\begin{tabular}{lcc}
\toprule
\multicolumn{1}{c}{\multirow{2}{*}{\textbf{Model}}} & \multicolumn{2}{c}{\textbf{Answer Accuracy (\%)}}
\\
\cmidrule(l){2-3} 
\multicolumn{1}{c}{}                                 & \textbf{PerfectIR} & \textbf{WholeDB} \\ \midrule
NeuralSPJ + Aggr (ours) &\textbf{90.10 $\pm$ 0.3} & - \\
T5 & 85.59 $\pm$ 0.2 & 65.96 $\pm$ 0.5  \\
Longformer & 76.43 $\pm$ 3 & 58.58 $\pm$ 0.4  \\
Fusion in Decoder & 39.61 $\pm$ 0.2 & 23.18 $\pm$ 0.6 \\
\bottomrule
\end{tabular}
\caption{T5 and Longformer both capture inter-fact attention whereas Fusion in Decoder does not. Regardless of how attention is used, using all facts in the database harms the model.}
\label{tab:baseline}
\end{table}

\begin{figure}[]
    \centering
    \includegraphics[width=\linewidth]{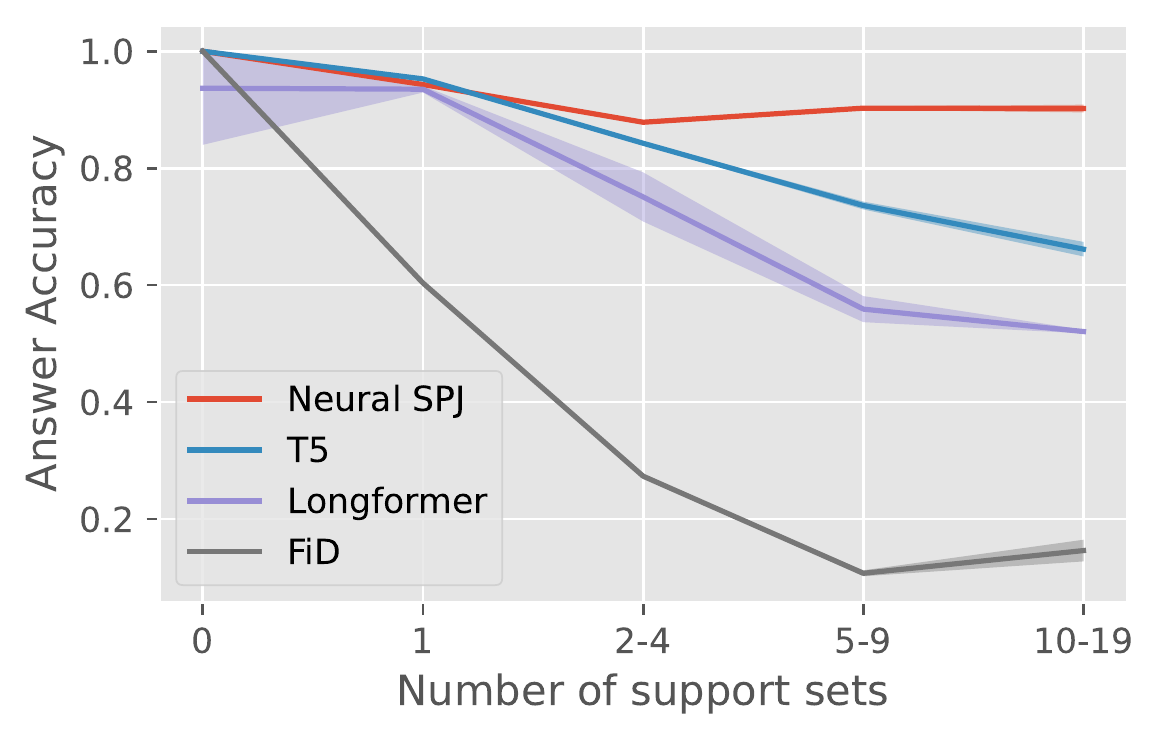}
    
    \caption{(PerfectIR) Even when provided with the correct contexts, baseline scores decrease for queries requiring the combination of multiple support sets.} 

    \label{fig:baseline}
\end{figure}



The overall scores, in Table~\ref{tab:baseline}, indicate that without a retrieval mechanism (i.e., WholeDB), all models were susceptible to distractor facts.
Furthermore, encoding all facts in a single model is not a viable solution to answer queries posed to \ndb s as this approach does not accurately answer queries that combine multiple support sets, illustrated in Figure~\ref{fig:baseline}, and cannot easily scale to thousands of facts.
Using a transformer yields errors when the query requires computation, such as counting, highlighted when comparing rows 1 and 3 of Table~\ref{table:pipeline}.

\paragraph{Inter-fact attention}
Applying FiD, which does not capture inter-fact attention, to scale to larger databases would not be successful because answer accuracy further decreases with with support set size. 
Applying Longformer, which captures inter-fact attention within a window could yield outcomes similar to the T5 transformer baseline where relevant facts are encoded with similar locality. However, in the limit, where context falls between different attention windows, the model could degrade to be similar to FiD.

\subsection{Evaluating the SSG+SPJ architecture}
Our architecture consists of a support set generator (SSG), a select-project-join (SPJ) operator that generates derivations over the support sets and an aggregation function over the results of the SPJ operators. 
Assuming a perfect SSG, the SPJ accurately answers more queries than the T5 transformer baseline (Table~\ref{tab:baseline}) because of the computation within the aggregation function that yields higher scores for min/max and count queries, displayed in Table~\ref{table:pipeline}. 
In combination with SSG, the overall score decreases to 67\% due to retrieval errors. However, SSG+SPJ still exceeds the WholeDB baselines.

It is tricky to evaluate the SSG in isolation because errors here not necessarily translate into errors in query answers. For example, the SSG may return a superset of a support set, but the SPJ may still generate the correct answer. 
Table~\ref{tab:supervised_issg} shows the performance of the SSG for a database of 25 facts. An output is considered an exact match if it is exactly the same as a support set in the reference data and soft match if it is a superset thereof.



\begin{table}[h]
\centering
\small

\begin{tabular}{lcccc}
\toprule
\multicolumn{1}{c}{\multirow{2}{*}{\textbf{Method}}} & \multicolumn{4}{c}{\textbf{Answer Accuracy (\%)}}                                                  \\
\cmidrule(l){2-5} 
\multicolumn{1}{c}{}                                 & \textbf{Min/Max} & \textbf{Bool} & \textbf{Count}  & \textbf{Set} \\ \midrule
SPJ PerfectIR & \textbf{89.72} & 99.10 & \textbf{94.68} & 85.25   \\
SSG + SPJ & 74.03 & 77.79 & 50.75 & 65.32  \\ \midrule 
\textit{T5 PerfectIR} & 78.23 & \textbf{99.34} & 87.33 & \textbf{89.19}\\

\bottomrule
\end{tabular}
\caption{Using retrieved evidence achieves results competitive to the PerfectIR on a DB of 25 facts.}
\label{table:pipeline}
\end{table}

\begin{table}[h!]
\small

\label{supervised_issg}

\centering
\label{ds_issg}
\label{table:ssg}
\begin{tabular}{@{}lcccc@{}}
\toprule
\multicolumn{1}{c}{\multirow{2}{*}{\textbf{\begin{tabular}[c]{@{}c@{}}Query \\ Type\end{tabular}}}} & \multicolumn{2}{c}{\textbf{\begin{tabular}[c]{@{}c@{}}Exact Match (\%)\end{tabular}}} & \multicolumn{2}{c}{\textbf{\begin{tabular}[c]{@{}c@{}}Soft Match (\%)\end{tabular}}} \\ \cmidrule(l){2-5} 
\multicolumn{1}{c}{}                                                                                & \multicolumn{1}{l}{\textbf{Precision}}          & \multicolumn{1}{l}{\textbf{Recall}}         & \textbf{Precision}                                & \textbf{Recall}                               \\ \midrule
Boolean                                                                                       & 64.00                                           & 80.28                                      & 66.15                                            &  80.68                                         
                                    \\

Set                                                                                                 &63.28                                          &  80.77                                      &65.23                                             & 81.30                                         \\
Count                                                                                               & 60.21                                           & 83.11                                       & 61.58                                            & 83.41                                         \\
Min/Max                                                                                        & 70.88                                            & 93.25                                        & 71.80                                           & 93.41        
                                \\ \midrule
                                \emph{Average}                                                                                                 & 65.96                                           & 86.51                                      & 67.36                                            & 86.82     
\\ 
\bottomrule
\end{tabular}
\caption{Precision and recall of supervised SSG w.r.t.\ the reference set. Note that errors in retrieval do not necessarily translate to wrong query answers because the SPJ operator is trained to be robust to noise. }
\label{tab:supervised_issg}
\end{table}

\paragraph{Decoding machine-readable outputs} The aggregation operator was selected by predicting a special token decoded by the SPJ. For 1.4\% of instances, an incorrect choice of aggregation function was made or the machine-readable outputs from the SPJ could not be parsed.


\subsection{Scaling to larger databases}
\begin{figure}[]
    \centering
    \includegraphics[width=\linewidth]{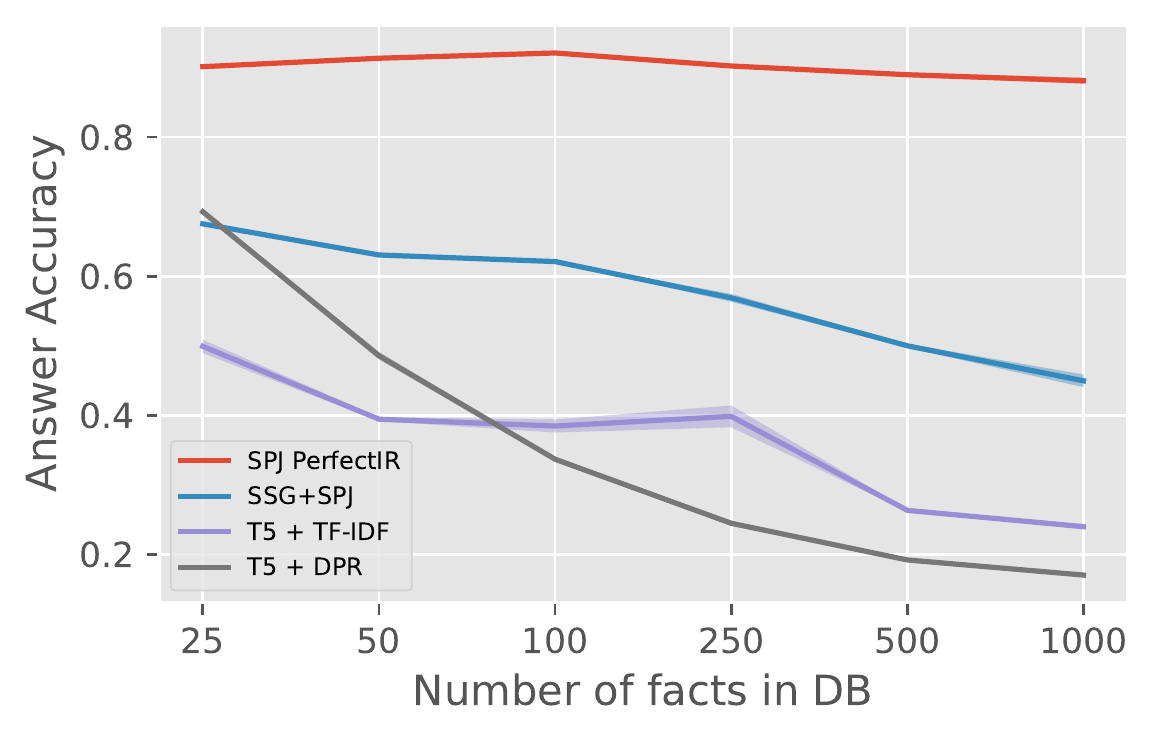}
    
    \caption{Scaling to larger databases with a model trained using 25 facts and tested on larger databases. }

    \label{fig:e2e}
\end{figure}

We scale the baseline transformers to larger databases using TF-IDF and DPR to retrieve appropriate facts. However, these models are still limited by the encoder size of the transformer.
In contrast, the SPJ operates over support sets of 1-2 facts and, in combination with the SSG, can scale to arbitrarily large databases, illustrated in Figure~\ref{fig:e2e}.
For Boolean queries, the combination of T5 and TF-IDF scored 89\%, exceeding the accuracy of the SSG+SPJ. This is because TF-IDF exploits token matching between the query and facts.  
For larger databases, the retrieval errors resulted in lower answer accuracy. While, with a perfect SSG, the the SPJ accurately answers most query types, as database size increases, the propagation of errors from the SSG resulted in erroneous answers.


\section{Related Work}



Database queries require reasoning over a large set of relevant and non-redundant facts and performing aggregation. 
While in-roads have been made to perform discrete reasoning and computation over passages \cite{Dua2019}, with explicit computation \cite{Andor2019} or differentiable modules \citep{nmn:iclr20}, these use only a single passage rather than requiring aggregation over large numbers of facts from different texts.

Multi-hop question answering requires finding supporting evidence in multiple documents (see \cite{welbl2018constructing,talmor2018web,Wolfson2020} for datasets facilitating this research). In answering multi-hop questions, the works decompose the question into simpler sub questions~\cite{min2019multi,Wolfson2020}, or condition each hop on the previously retrieved documents~\cite{asai2019learning,xiong2020answering}.



While tasks such as ComplexWebQuestions~\cite{talmor2018web} and BREAK~\cite{Wolfson2020} focus on complex queries that can be broken down into simpler ones, our focus is on set-based and aggregation queries where the complexity comes from the need to retrieve and process a large number of non-redundant relevant facts. In contrast to the \emph{set} and \emph{count} tasks in bAbI~\cite{weston2015towards}, where each query is based on a small context (less than 20 facts), our dataset scales from databases of 25 facts to 1000.

Bridging the gap between unstructured natural language data and database-style querying has been a long-standing theme in database research~\cite{DBLP:conf/cidr/HalevyED03}. 
The work on information extraction has developed techniques for translating segments of natural language text into triples that can be further processed by a database system. 
There has been significant work on translating queries posed in natural language into SQL queries on a database whose schema is known~\cite{Androutsopoulos1995, Li2014,DBLP:journals/corr/abs-2007-15280}, with extensions to semi-structured data and knowledge bases~\cite{Pasupat2015,Berant2013}. More recently, systems such as {BREAK}~\cite{Wolfson2020} and {ShARC}~\cite{saeidi2018interpretation} have trained models to translate a natural language query into a sequence of relational operators (or variants thereof).

\section{Conclusions}

Database systems are the workhorse of data analysis but they require a pre-defined schema. Part of their power stems from the fact that a data analyst can explore the data by easily posing a wide variety of queries. Given the rise in the amount of data that is becoming available in text, images and other modalities, we would like to build systems that enable the flexibility of posing complex queries against such data,  but without the need for a pre-defined schema. 

This paper proposed an architecture for neural databases and the associated \wikidb\ dataset, as first steps towards realizing a system for querying multi-modal data. Our architecture is capable of overcoming the limitations of transformer models because it runs multiple transformers in parallel, each taking a small set of facts. Consequently, \ndb s can scale to large databases.

Additional research is required in order to scale \ndb s to larger datasets, more complex queries,  and to multi-modal data. In particular, one of the key components of the architecture is the SSG module that retrieves the relevant facts to feed to each instance of the neural SPJ. We believe that in practice, the semantics of the application will provide a strong hint on which facts may be relevant. For example, when querying a large corpus of social-media posts, each post is a candidate support set as long as the query does not require joining data from multiple posts. In addition, we assumed that our databases describe a snapshot of the world. In practice, we may have facts that override previous ones (e.g., `Samantha works for Apple', followed by `Samantha works for Twitter') and we would need to reason about which facts should be ignored. 




\section*{Acknowledgments}
We would like to thank Yann LeCun and Antoine Bordes for the initial discussion that sparked the idea of neural databases. 
This work was performed while James Thorne  and Fabrizio Silvestri were at Facebook.

\section*{Broader Impact Statement}
\paragraph{Ethical Concerns}
A NL database is very similar to a traditional database in terms of applications with a difference that it extends the use of databases on unstructured text. For example, NL databases can be used to produce analytics on data expressed in natural language. For an NL database to be applicable in the context of a virtual assistance, they will likely need to be trained on real-world conversations. Privacy preserving ML methods should be considered for such applications.

\paragraph{Environmental Concerns}
Large transformer-based models take a lot of computational resources and energy for pre-training and fine-tuning. As a result such models raise environmental concerns. In our proposed architecture, we only fine-tune transformer models on small support sets. We then use several instances of such models in parallel for inference, instead of a single large model, even on large datasets. Therefore, the model is relatively efficient, both during the fine-tuning and during the inference.

\bibliographystyle{acl_natbib}
\bibliography{references}

\clearpage 
\onecolumn 
\appendix
\section{Appendix}
\subsection{Sample Data and Dataset Statistics}
\label{appendix:stats}
\begin{table*}[h]
\small
\centering
\begin{tabular}{p{3cm} p{12cm}}
\hline
\multicolumn{2}{l}{Example: \texttt{Set}} \\
\hline
Question     &  Who studied at University of Minnesota?\\
\\
Supporting Facts     &  \textbf{1. } [John B Totushek was born on 7 September 1944 in Minneapolis. He attended the University of Minnesota and became a US Naval Aviator. Mr. Totushek was also a human being.] \newline \textbf{2. } [Melvin Maas graduated from the University of Minnesota and is buried at Arlington National Cemetery. He is a native of Minnesota and his language is English.] \newline \textbf{3. } [Clarence Larson graduated from the University of Minnesota and is a member of the National Academy of Engineering.] \newline \textbf{4. } [Ted Mann, who is the surname of Ted Mann, attended Duke University and the University of Minnesota. He is a human being.]\\
\\
Answer &  [John B. Totushek, Ted Mann, Clarence Larson, Melvin Maas]\\
\hline
\multicolumn{2}{l}{Example: \texttt{count}} \\
\hline
Question     &  How many people work for Yale Law School?\\
\\
Supporting Facts     & \textbf{1.} [Michael Ponsor, born in Oxford, graduated from Pembroke College in Oxford. He was awarded the Rhodes Scholarship and is an employee at Yale Law School. He is an expert in the field of human rights.] \newline \textbf{2.} [Stephen Wizner is an American legal scholar who graduated from Dartmouth College and is a graduate of the University of Chicago Law School. He works at Yale Law School.]\\
\\
Answer & 2 \\
\hline
\multicolumn{2}{l}{Example: \texttt{Min/Max}} \\
\hline
Question     &  What is the largest yearly attendance?\\
\\
Supporting Facts     &  \textbf{1. } [The musee en herbe has a visitor per year of] 70000. \newline
                        \textbf{2. } [The total number of visitors to the Hirschsprung Collection is 71779 per year.] \newline
                        ... \newline
                        \textbf{24.} [The Tate Modern has a visitor count of 5839197 visitors per year.] \newline
                        \textbf{25.} [Catoctin Mountain Park attracts 221750 visitors per year.] \newline
\\
\\
Answer &  5839197\\
\hline
\multicolumn{2}{l}{Example: \texttt{Bool}} \\
\hline
Question     &  Is North Carolina State University the employer of Wes Moore?\\
\\
Supporting Facts     &  \textbf{1. } [Wes Moore is a human being who is employed at Francis Marion University and is a basketball player for North Carolina State University.]\\
\\
Answer &  TRUE\\
\hline
\multicolumn{2}{l}{Example: \texttt{Join}} \\
\hline
Question     &  Who plays for a team in Ligue 1?\\
\\
Supporting Facts     &  \textbf{1. } [Thomas Allofs started his career in 1989 with RC Strasbourg Alsace. He finished his career in 1990., \newline RC Strasbourg Alsace is an association football club in the Ligue 1 league. It was founded in 1906 and is located in Strasbourg, France.]\\
\\
Answer &  [Thomas Allofs]\\
\hline
\end{tabular}
\caption{Examples of different types of queries, their supporting facts and answers. These examples are based on databases of size 25.}
\label{tab:db-examples}
\end{table*}

\begin{figure}
    \centering
    \begin{subfigure}[b]{0.35\textwidth}
        \includegraphics[width=\textwidth]{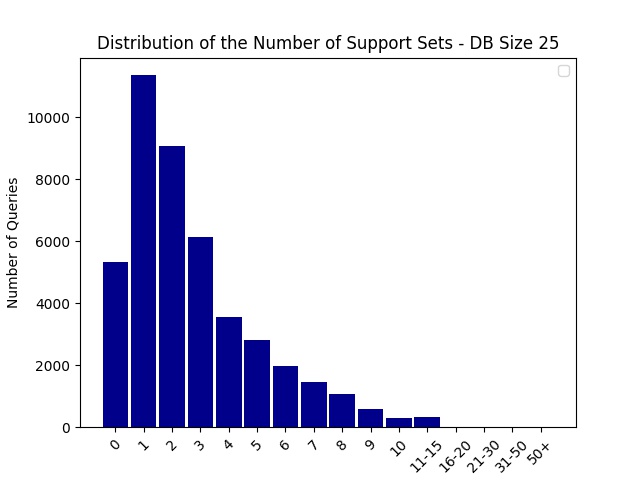}
        \label{fig:hist_200}
    \end{subfigure}
    ~ 
    \begin{subfigure}[b]{0.35\textwidth}
        \includegraphics[width=\textwidth]{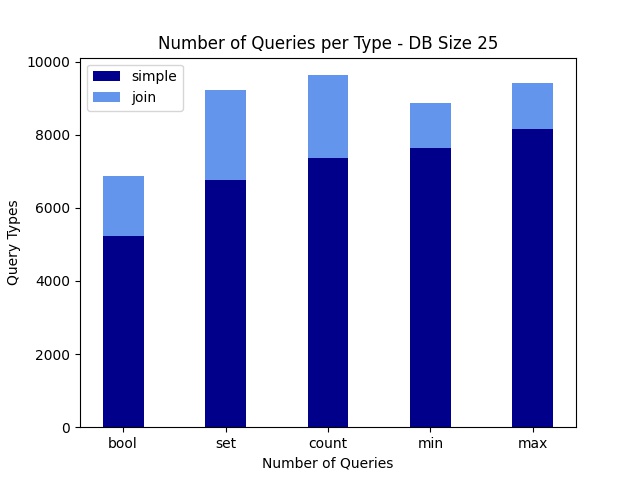}
        \label{fig:qtype_100}
    \end{subfigure}

    \begin{subfigure}[b]{0.35\textwidth}
        \includegraphics[width=\textwidth]{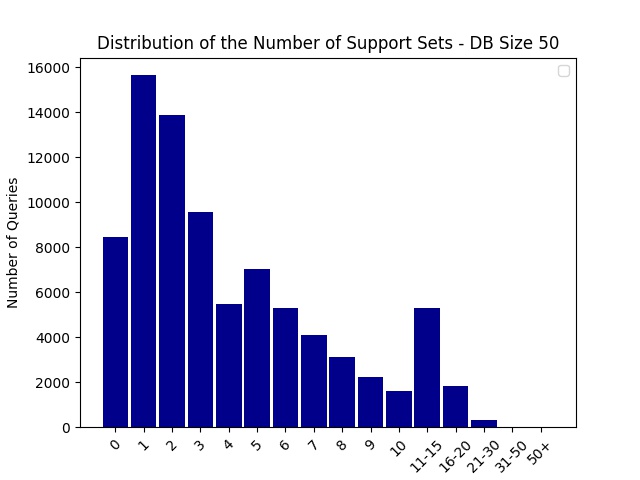}
        \label{fig:hist_200}
    \end{subfigure}
    ~ 
    \begin{subfigure}[b]{0.35\textwidth}
        \includegraphics[width=\textwidth]{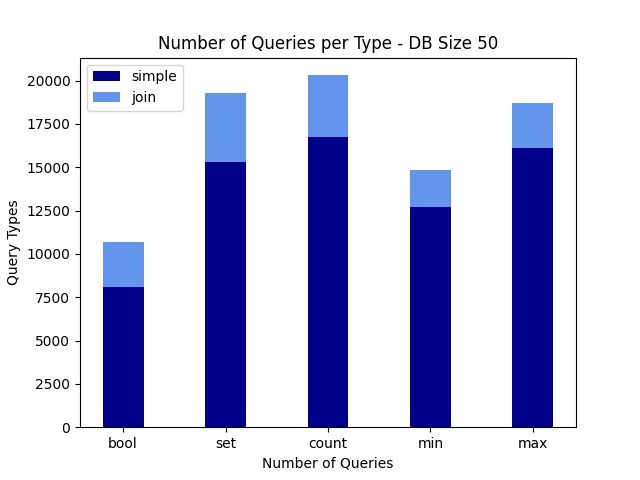}
        \label{fig:qtype_100}
    \end{subfigure}
    \begin{subfigure}[b]{0.35\textwidth}
        \includegraphics[width=\textwidth]{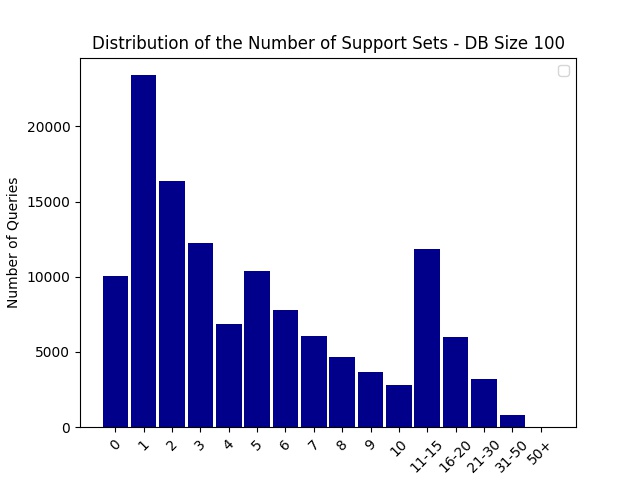}
        \label{fig:hist_250}
    \end{subfigure}
    ~ 
    \begin{subfigure}[b]{0.35\textwidth}
        \includegraphics[width=\textwidth]{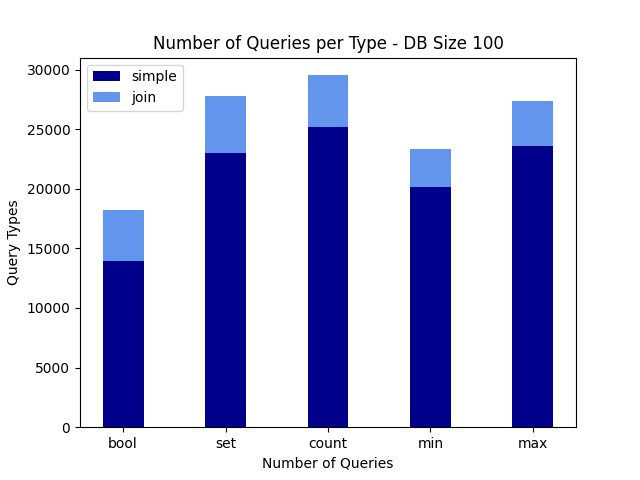}
        \label{fig:qtype_250}
    \end{subfigure}
    
    \begin{subfigure}[b]{0.35\textwidth}
        \includegraphics[width=\textwidth]{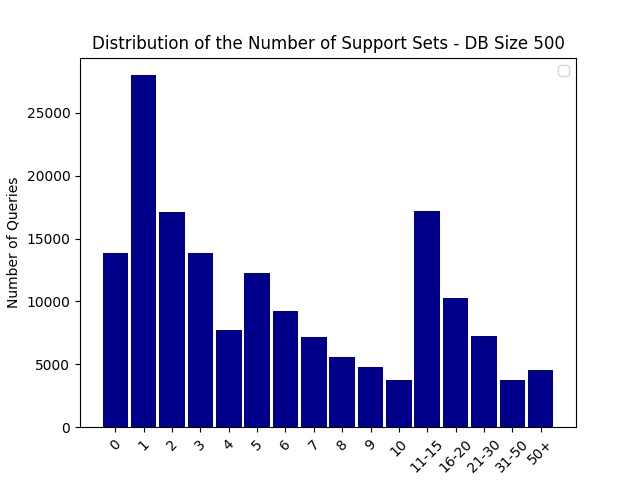}
        \label{fig:hist_1000}
    \end{subfigure}
    ~ 
    \begin{subfigure}[b]{0.35\textwidth}
        \includegraphics[width=\textwidth]{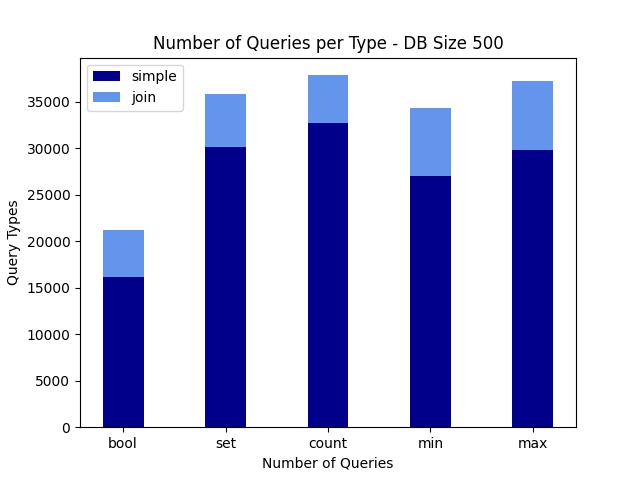}
        \label{fig:qtype_1000}
    \end{subfigure}

    \begin{subfigure}[b]{0.35\textwidth}
        \includegraphics[width=\textwidth]{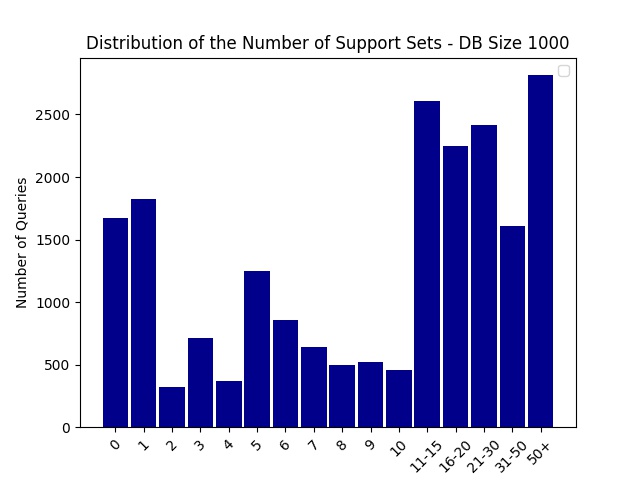}
        \label{fig:hist_2500}
    \end{subfigure}
    ~ 
    \begin{subfigure}[b]{0.35\textwidth}
        \includegraphics[width=\textwidth]{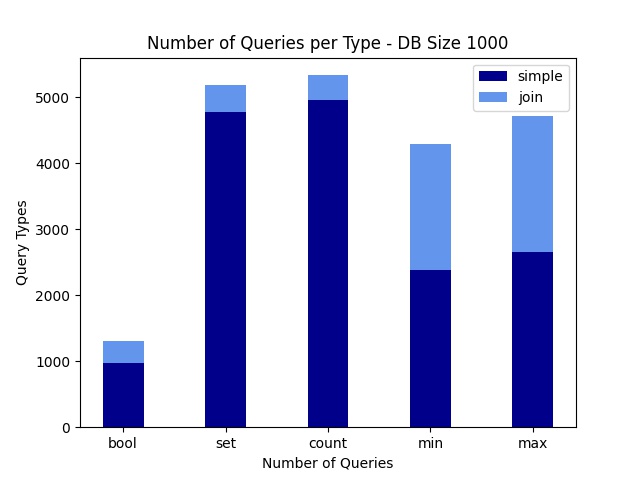}
        \label{fig:qtype_2500}
    \end{subfigure}
    \caption{Dataset statistics for DBs of varying sizes provided with \wikidb}
    \label{fig:stat_all}
\end{figure}

\end{document}